\def\year{2020}\relax
\documentclass[letterpaper]{article} 
\usepackage{aaai20}  
\usepackage{times}  
\usepackage{helvet} 
\usepackage{courier}  
\usepackage[hyphens]{url}  
\usepackage{graphicx} 
\usepackage{graphicx}  
\usepackage[ruled]{algorithm2e}
\usepackage{subcaption}
\usepackage{multirow}
\usepackage{pifont}
\newcommand{\cmark}{\ding{51}}%
\newcommand{\xmark}{\ding{55}}%

\title{Efficient Facial Feature Learning with Wide Ensemble-based\\Convolutional Neural Networks}
\author{Henrique Siqueira, Sven Magg and Stefan Wermter\\
Knowledge Technology\\
Department of Informatics, University of Hamburg\\
Vogt-Koelln-Str. 30, 22527 Hamburg, Germany\\
\{siqueira, magg, wermter\}@informatik.uni-hamburg.de 
}

\begin{document}

\maketitle

\begin{abstract}
Ensemble methods, traditionally built with independently trained de-correlated models, have proven to be efficient methods for reducing the remaining residual generalization error, which results in robust and accurate methods for real-world applications. In the context of deep learning, however, training an ensemble of deep networks is costly and generates high redundancy which is inefficient. In this paper, we present experiments on Ensembles with Shared Representations (ESRs) based on convolutional networks to demonstrate, quantitatively and qualitatively, their data processing efficiency and scalability to large-scale datasets of facial expressions. We show that redundancy and computational load can be dramatically reduced by varying the branching level of the ESR without loss of diversity and generalization power, which are both important for ensemble performance. Experiments on large-scale datasets suggest that ESRs reduce the remaining residual generalization error on the AffectNet and FER+ datasets, reach human-level performance, and outperform state-of-the-art methods on facial expression recognition in the wild using emotion and affect concepts.
\end{abstract}

\section{Introduction}
``We get resourcefulness from having many resources; not from having one very smart one'' \cite{Minsky2019}. In machine learning, ensemble methods refer to a set of models where an inference is made collectively based on individual predictions \cite{Dietterich2000}. A well-trained ensemble can reduce the remaining residual generalization error, which results in predictions being more accurate than any single model in the ensemble. Traditional ensemble (TE) methods are built by independently training several models on the same or different data. They can be composed of a single type of machine learning method such as an ensemble of neural networks \cite{Hansen1990}, but the diversity is often higher when an ensemble is built from a library of different methods \cite{Caruana2004}.

At present, ensembling of deep networks is an important resource but requires high computational power. To make this training-intensive technology accessible to everyone, recent studies have explored ways to reduce redundancy in ensembling. \citeauthor{Meshgi2018} (\citeyear{Meshgi2018}) have exploited concepts from active learning to reduce training time and redundancy. Rather than using the whole dataset for training, their ensemble method is trained on the most informative samples that maximize learning based on the query by committee algorithm \cite{Seung1992}.

Another approach adopted a divide-and-conquer strategy \cite{Li2019} where the input space is decomposed into multiple regions, and each region is used to train one convolutional neural network of the ensemble. Despite their progress on reducing redundancy, their approaches fall within the ``explicit'' ensemble methods, i.e., consist of independent models. Therefore, redundancy of low-level visual features is still high, and unnecessary computational resources have to be allocated for processing such features.

In the so-called ``implicit'' ensemble methods, a single network may generalize as well as an ensemble by distilling its knowledge \cite{Hinton2015}. By training a convolutional neural network (CNN) with the outputs of an ensemble of CNNs, \citeauthor{Shen2019} (\citeyear{Shen2019}) have reduced inference time and redundancy while maintaining generalization power and similar intermediate representations under an adversarial training strategy. However, training time is greatly increased with their approach since a trained traditional ensemble is a fundamental pre-requisite.

\begin{figure}[!t]
	\centering
	\includegraphics[width=.47\textwidth]{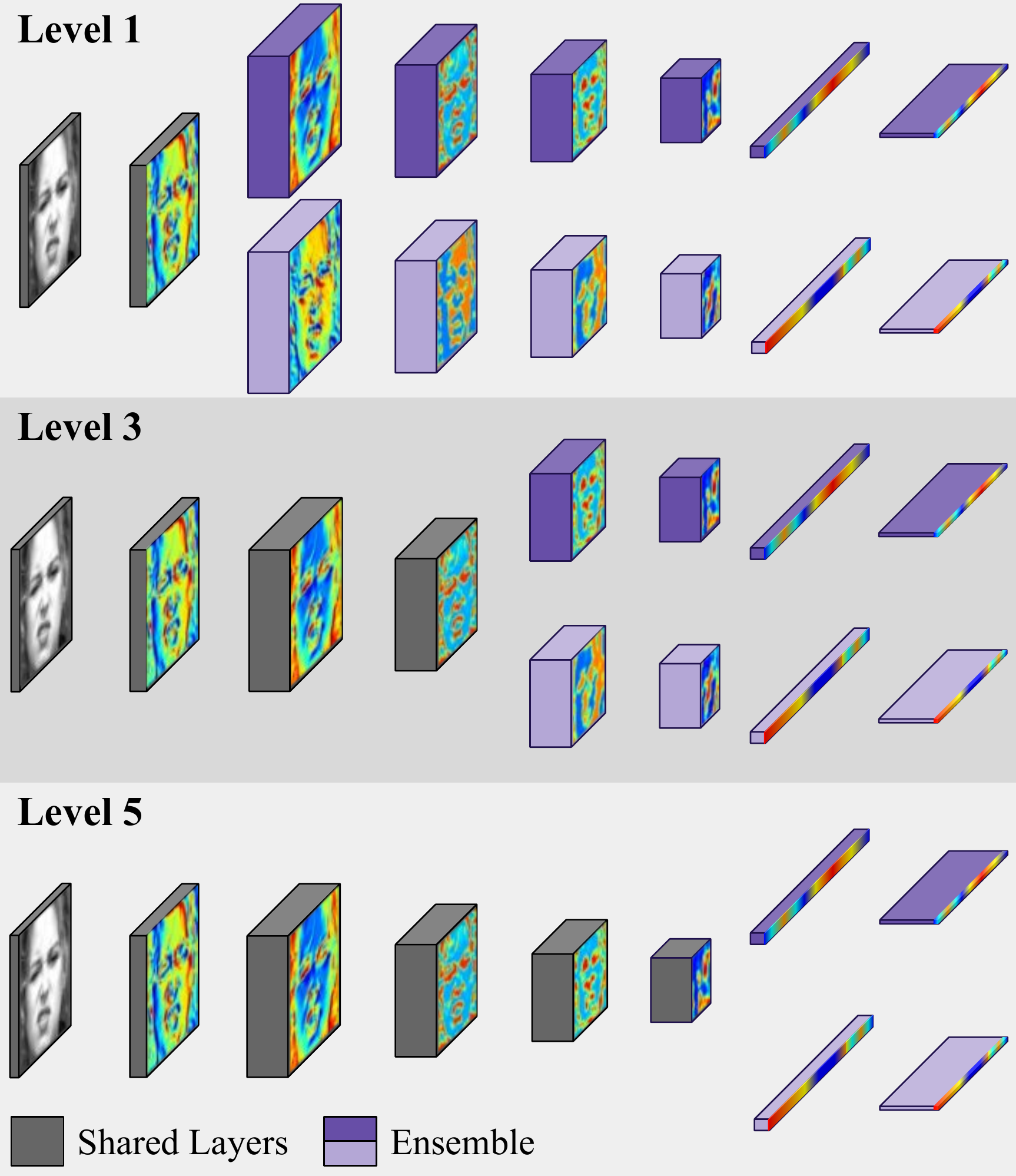}
	\caption{Ensemble with Shared Representations (ESR). Illustration of the experiments to investigate the effect of branching level on computational load and generalization.}
	\label{fig:illustration}
\end{figure}

Ensemble with Shared Representations (ESR), proposed in our previous work \cite{Siqueira2018a}, offers the best of the two worlds. It is neither a fully implicit nor a fully explicit ensemble method. As depicted in Figure \ref{fig:illustration}, the shared layers represent the implicit part. They are responsible for the reduction of redundancy, training, and inference time. The low-level features learned by them are shared with the ensemble of convolutional branches. The latter characterizes the explicit part and carries the diversity of the ensemble. The level to start the ensemble of branches plays an important role in the computational load and generalization power as well as for redundancy and diversity. However, the effect of the branching level is still an open question that needs careful analysis. In the context of facial expression recognition, for instance, starting branching too early (level 1) may result in high redundancy of low-level facial features where all branches have to learn skin textures and so forth. On the other hand, branching too late may drastically decrease diversity in the ensemble where features from the shared layers no longer correspond to spatial facial features (level 5). We hypothesize that the optimal branching level may be located between the extremes, where the abstraction level of the facial features is high including smiling and frowning but have yet to be encoded into emotion concepts.

Another aspect that needs further understanding is the scalability of ESRs to large-scale datasets of facial expressions of emotion. Can ESRs reduce the remaining residual generalization error when training data is abundant? After reviewing prior work on facial expression recognition, we address these questions. In this paper, the effects of varying the branching level are extensively examined, quantitatively and qualitatively, first, on a small-scale but clean and well-structured dataset of facial expressions \textit{in the lab}. Subsequently, experiments using a single GeForce GTX 1080 on large-scale benchmarks for facial expression recognition \textit{in the wild} demonstrate the affordability and scalability of ESRs, followed by conclusions and future research. For reproducibility purposes, source code of our experiments, the ESR implementation in PyTorch, trained networks and supplementary material are available in our GitHub repository\footnote{Source code: https://github.com/knowledgetechnologyuhh/\\Efficient-Facial-Feature-Learning-with-Wide-Ensemble-based-Convolutional-Neural-Networks}.

\subsection{Prior Work on Facial Expression Recognition}
Early approaches for automatic facial expression recognition have followed the general pipeline to tackle computer vision problems, which consist of pre-processing the facial images, appearance and/or geometric hand-crafted feature extraction and, in the final stage, the classification of such features \cite{Tian2005}. These methods are usually fast and accurate in indoor environments but frequently drop in performance under real-world conditions \cite{Mollahosseini2019}.

The rapid progress in deep learning motivated researchers to develop facial expression recognition systems using deep neural networks. Since these networks can automatically learn features from data, hand-feature engineering was left out in the pipeline. Besides that, feature learning allows deep networks to learn a broader range of facial features than earlier approaches, including variation to rotations, and illumination changes. Indeed, as investigated by \citeauthor{Khorrami2015} (\citeyear{Khorrami2015}), it has turned out that the features learned by a CNN trained for facial expression recognition reflect the facial features of emotion suggested by the psychologist Paul Ekman during his study of universal facial expressions of emotion \cite{Ekman1989}. Recent approaches rely on well-established networks for object recognition such as AlexNet, MobileNet, ResNet, and VGGNet \cite{Barsoum2016,Hewitt2018,Mollahosseini2019}. In visual perceptual tasks, certain features previously learned can be transferred among related tasks and the use of pre-trained networks often speed up learning and culminate in better accuracy than training them from scratch. These approaches represent the state of the art in the datasets utilized in our experiments (for a review, see \cite{Poria2017}).

\section{Ensembles with Shared Representations}
Ensembles with shared representations exploit the fundamental properties of convolutional networks. A convolutional layer learns \textit{local patterns} from the previous layer by convolving small filters over its input feature space \cite{Chollet2018}. Thus, the patterns learned by convolutional layers are translation-invariant. Another property is the capability to learn spatial hierarchies of patterns by stacking multiple convolutional layers. Consider the intermediate representations exhibited in Figure \ref{fig:illustration}. Early layers learn simple and local visual patterns such as oriented lines, edges, and colors. These low-level abstractions of input space are the reason for early feature maps resembling a face with emphasis on certain regions. Subsequent layers hierarchically combine local patterns from previous layers into increasingly complex concepts such as nose, mouth, and eyes. The level of abstraction increases as you go deeper into the network until the point where feature maps are no longer visually interpretable. Finally, the last layer encodes these representations into semantic concepts, for instance, concepts of emotion.

These properties are the foundations of ESRs and play a crucial role in reducing redundancy of visual features in the ensemble. An ESR consists of two building blocks. (1) The base of the network (gray blocks in Figure \ref{fig:illustration}) is an array of convolutional layers for low- and middle-level feature learning. (2) These informative features are then shared with independent convolutional branches (purple blocks) that constitute the ensemble. From this point, each branch can learn distinctive features while competing for a common resource - the shared layers. This competitive training emerges from the minimization of a \textit{combined loss function} defined as the summation of the loss functions of each branch as follows:
\begin{equation}
L_{esr} = \sum_{b} \sum_{i} L[P(f(x_{i}) = y_{i} | x_{i}, \theta_{shared}, \theta_{b}), y_{i}],
\label{eq:combined_loss}
\end{equation}
where $b$ denotes the branch index, $(x_{i}, y_{i})$ random samples from the training set, $\theta_{shared}$ the parameters of the shared layers from the base of the network that acts as a regularizer for ESRs, and $\theta_{b}$ the parameters of a convolutional branch that composes the ensemble.

Because novel convolutional branches are added in sequence while training, as outlined in Algorithm 1, the shared layers turn out to be an efficient transfer learning mechanism that guides and accelerates learning as the ensemble grows. Besides that, the shared representations are conditioned to learn features that are suitable to different branches in the ensemble due to the inductive transfer learning from the combination of multiple loss functions from each convolutional branch. During inference time, a given input is classified by the ensemble through a collective decision such as plurality and majority voting.

\begin{algorithm}[h]
	\SetAlgoLined
	initialize the shared layers with $\theta_{shared}$\\
	\For{b \KwTo maximum ensemble size}{
		initialize the convolutional branch $B_{b}$ with $\theta_{b}$\\
		add the branch $B_{b}$ to the network $ESR$\\
		sample a subset $D'$ from a training set $D$\\
		\ForEach{mini-batch $(x_{i}, y_{i}) \sim D'$}{
			perform the forward phase\\
			initialize the combined loss function $L_{esr}$ to $0.0$\\			
			\ForEach{existing branch $B_{b'}$ in $ESR$}{
				compute the loss $L_{b'}$ with respect to $B_{b'}$\\
				add $L_{b'}$ to $L_{esr}$\\
			}
			perform the backward phase\\
			optimize $ESR$\\
		}
	}
	\caption{Training ESRs.}
\end{algorithm}

\section{Experimental Datasets}
Over the last two decades, a number of datasets of facial expressions have been collected for research on affective computing \cite{Mollahosseini2019}. Among the attributes that characterize these datasets (e.g. the number of subjects and representations of emotion), the nature of the facial expressions is critical for developing and assessing automatic facial expression recognition systems.

Some of the datasets rely on posed or simulated facial expressions of emotion. They are supported by Ekman and Friesen's work \cite{Ekman1976,Ekman1989} on universals in facial expressions of emotion. The arguments about universality suggest that when we feel certain emotions, some facial movements manifest regardless of age, culture, race, or sex. For example, when you are angry in a traffic jam, you scowl; when you are happy after an acceptance notification, you smile. These visible facial movements have been mapped latter to the Facial Action Code (FAC) \cite{Ekman1976}, where every single appearance change (action unit, AU) was categorized. These datasets are occasionally called \textit{in-the-lab} datasets. As the name states, facial images are collected in controlled indoor environments where experimental variables (e.g. scene lighting and camera-view points) are accurately adjusted. They usually provide clean and high-quality data. Although posed emotional expressions are considered more expressive than natural expressions in everyday life \cite{Koolagudi2012}, the datasets are well structured and carefully annotated from emotions to FAC \cite{Lucey2010}.

On the other end of the spectrum, there are the \textit{in-the-wild} datasets with spontaneous facial expressions. Over a century, since Charles Darwin published \textit{The Expression of the Emotions in Man and Animals} \cite{Darwin1872}, the universality of emotional expressions has been called into question by distinguished psychologists including William James \cite{James1884,James1890}, James A. Russell \cite{Russell2003} and Lisa F. Barrett \cite{Barrett2015,Barrett2017,Gendron2018}. Their theses converge to the same conclusion: diversity of emotional expressions is the norm, not the uniformity. According to \citeauthor{James1890} (\citeyear{James1890}), any categorization of an emotional expression can be seen ``as true and as `natural' as any other''. Nevertheless, Russell argues for the minimum universality in his core affect theory, where emotions are described in an orthogonal dimensional space of arousal and valence levels. Therefore, even though we cannot claim that \textit{in-the-wild} datasets contain emotional facial expressions, they do provide large and rich data of facial configurations captured in a vast range of environmental conditions. These variations are crucial to develop robust facial expression recognition systems. In most cases, the data is gathered from films or the Internet and annotated based on affect concepts and/or emotion concepts \cite{Mollahosseini2019}.

\begin{figure}[ht]
	\centering
	\includegraphics[width=.47\textwidth]{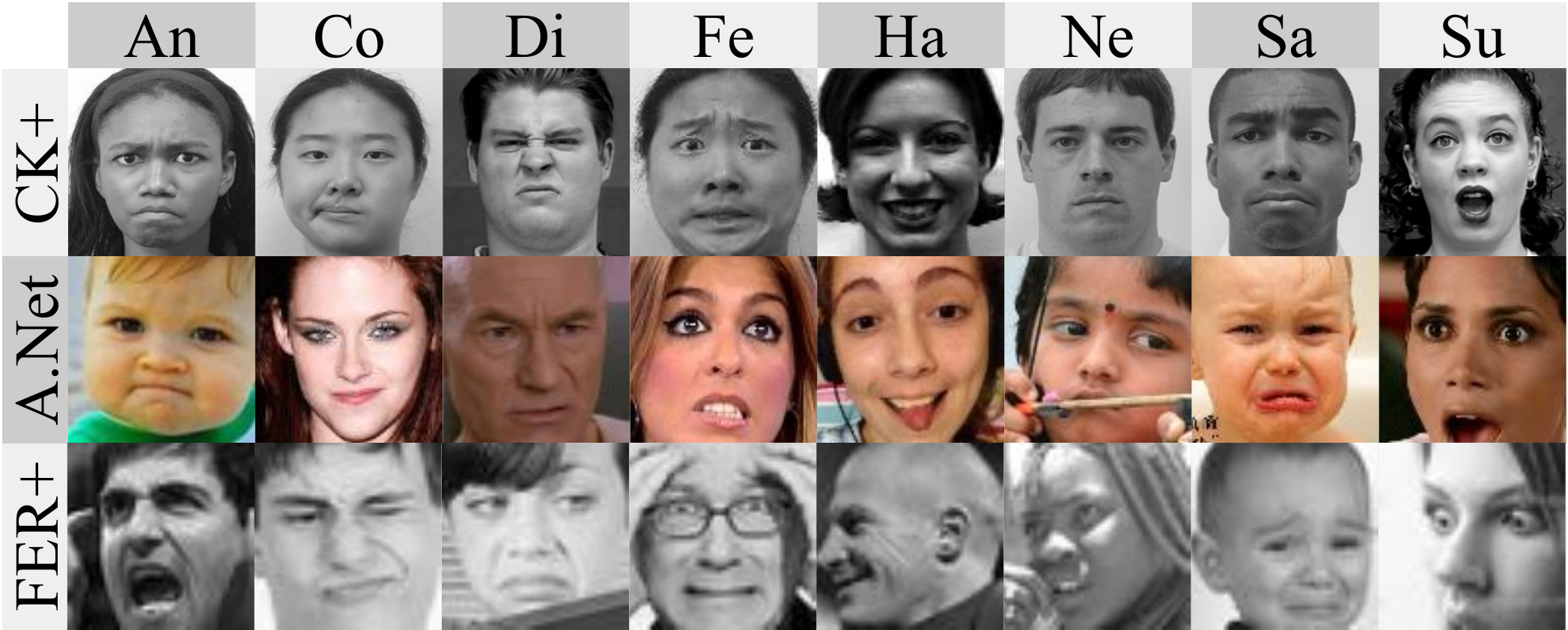}
	\caption{Experimental datasets. Extended Cohn-Kanade (CK+), AffectNet and FER+, from the top to the bottom.}
	\label{fig:datasetshort}
\end{figure}

We trained and tested the ensemble with shared representations on in-the-lab and in-the-wild datasets for a couple of reasons. The former allows us to evaluate ESRs' inference performance when training data is scarce and to conduct a descriptive analysis of their predictions based on the FAC system. On the other hand, the latter permits us to asses the scalability of ESRs to large-scale datasets and to test their inference performance in a more challenging scenario which includes, among other aspects, a vast intraclass variation, rotations, occlusions, and a heavily imbalanced label distribution. Together, they provide evidence on how flexible and robust ensembles with shared representations are in dealing with different shortcomings on facial expression recognition in the lab and in the wild. A few samples from the datasets used in our experiments are shown in Figure \ref{fig:datasetshort} and the technical details are described as follows.

\subsection{In-the-Lab Dataset}
\textbf{The Extended Cohn-Kanade (CK+) dataset} \cite{Lucey2010} has been vastly used to develop action unit detection and facial expression recognition systems. 123 subjects between 18 and 50 years old from different races, sex, and ethnic groups were told to portray a series of facial configurations based on FAC. The onset facial expressions were recorded from frontal and 30-degree camera-view points, and their peaks were carefully annotated and validated in terms of 30 action units and 8 discrete emotion concepts.

\subsection{In-the-Wild Datasets}
\textbf{AffectNet} \cite{Mollahosseini2019} is the largest dataset of facial expressions in the wild publicly available. It contains more than one million images retrieved from the Internet using emotion keywords from different languages, where half of them were manually annotated by human experts using 8 discrete emotions, arousal and valence levels. In addition to its heterogeneity, the heavily imbalanced label distribution (e.g., contempt constitutes only 1\% of the annotated images) and the strong baselines pose a real challenge for the affective computing community.

\textbf{FER+} \cite{Barsoum2016} derives from the re-annotation of the Facial Expression Recognition 2013 (FER-2013) dataset \cite{Goodfellow2015} due to the originally high degree of noise presented in the annotations. FER-2013 was created by querying facial images from Google's image search engine using 184 emotion keywords. Each of the 35,887 facial images was then re-labeled by 10 annotators using crowd-sourcing, and the contempt category was added to the dataset as one of the possible 8 emotion labels.

\section{Redundancy and Diversity Analysis}
\begin{figure*}[!t]
	\centering
	\includegraphics[width=1.0\textwidth]{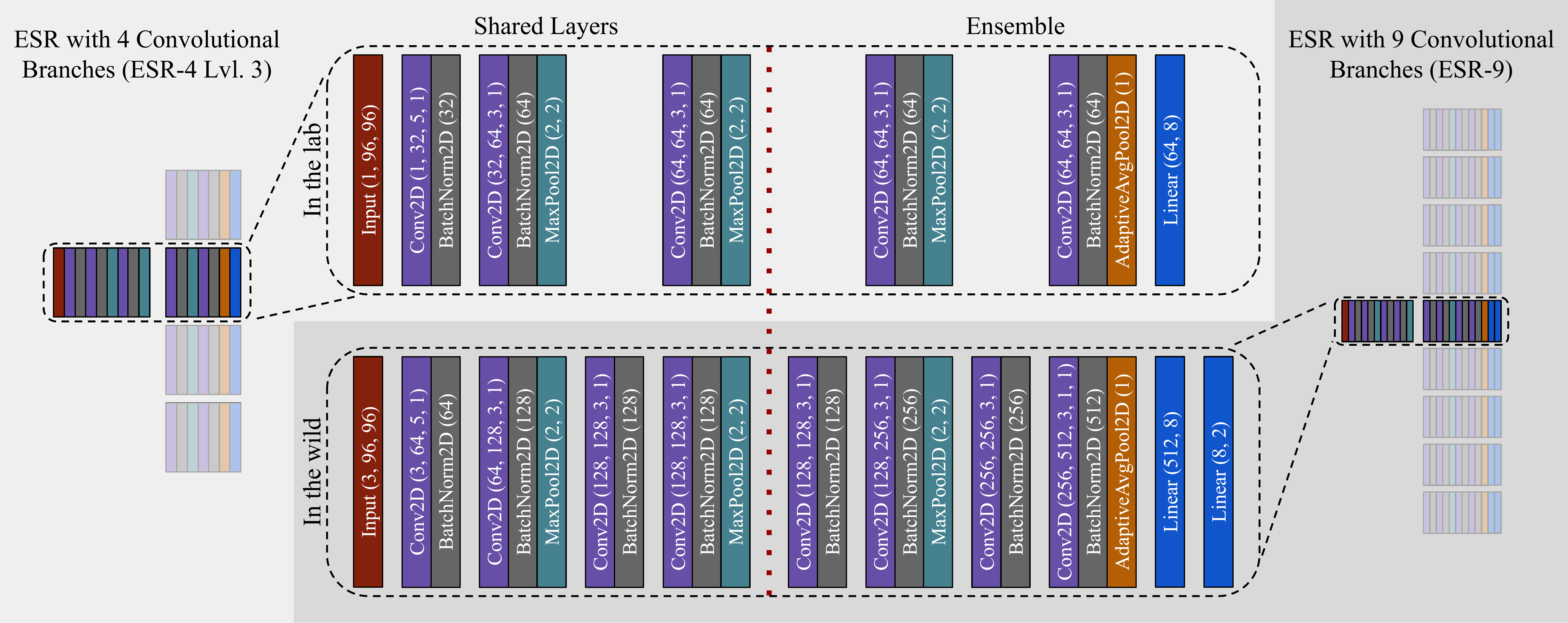}
	\caption{On the top, the architecture used in the in-the-lab experiments and an illustration of ESR-4 Lvl. 3 on the left. On the bottom, the architecture used in the in-the-wild experiments and an illustration of ESR-9 on the right. The latter architecture was designed to be equivalent to the former with respect to the spatial information of the features. The ReLU activation function is applied after batch normalization layers. The last linear layer in the bottom architecture was added for the affect perception experiment only. Each color represents a different type of layer and the PyTorch nomenclature was followed for reproducibility.}
	\label{fig:architecture}
\end{figure*}

We start this section describing the methodology adopted to explore the impact of the branching level on redundancy, and diversity of ESRs. After discussing training strategies and architectural design, we present quantitative results on computational load, redundancy and recognition performance. We conclude this section by presenting evidence that ESRs converge faster than a TE while preserving diversity, by analyzing convergence graphs and saliency maps via Grad-CAM \cite{Selvaraju2017} at different training milestones.

\subsection{Methodology}
We followed the subject-independent 10-fold cross-validation for comparison purposes based on our previous work \cite{Siqueira2018a}. First, we extracted the first and last three frames from each sequence on CK+, converted them to gray-scale, cropped the faces using the Viola and Jones's algorithm (\citeyear{Viola2004}), and resized the facial images to 96 x 96 pixels. The first frame was labeled as neutral, whereas the last three frames received one of the seven basic emotion labels. Subsequently, the images were separated into 10 folds according to the subject's id available in the metadata. Each fold was populated with facial images from a subject by iterating the subject id and the fold id, which resulted in 12 subjects and 130.8 facial images on average for each fold. With the folds populated, we run the experiment 10 times. In each trial $t$, we selected fold-$(t)$ for testing, fold-$(t+1)$ for validating, and only the first four folds from the remaining eight folds for training, i.e., 523.2 images on average on the training set.

\subsection{Training ESRs at Different Branching Levels}
How does the branching level affect computational load, redundancy and recognition performance on ESRs? This research question was addressed by training several ESRs at different branching levels and analyzing the impact on those aspects. Two baselines were defined according to our previous research \cite{Siqueira2018a}. After an exhaustive search among different convolutional architectures and training strategies, the network with the best mean test accuracy on CK+ was selected as the first baseline. The network comprises five convolutional layers, each followed by a batch normalization layer. A max-pooling layer was also added after the second, third, and fourth batch normalization layers. On top, a global average pooling layer transforms the last feature maps into a vector and forwards it to the dense output layer for facial expression recognition. The ReLU activation function was applied after batch normalization layers as suggested by \citeauthor{Ioffe2015} (\citeyear{Ioffe2015}). A detailed architectural diagram of the network used in our experiments is presented at the top of Figure \ref{fig:architecture}. The second baseline is a traditional ensemble with four of such networks.

\begin{figure}[!t]
	\centering
	\includegraphics[width=.47\textwidth]{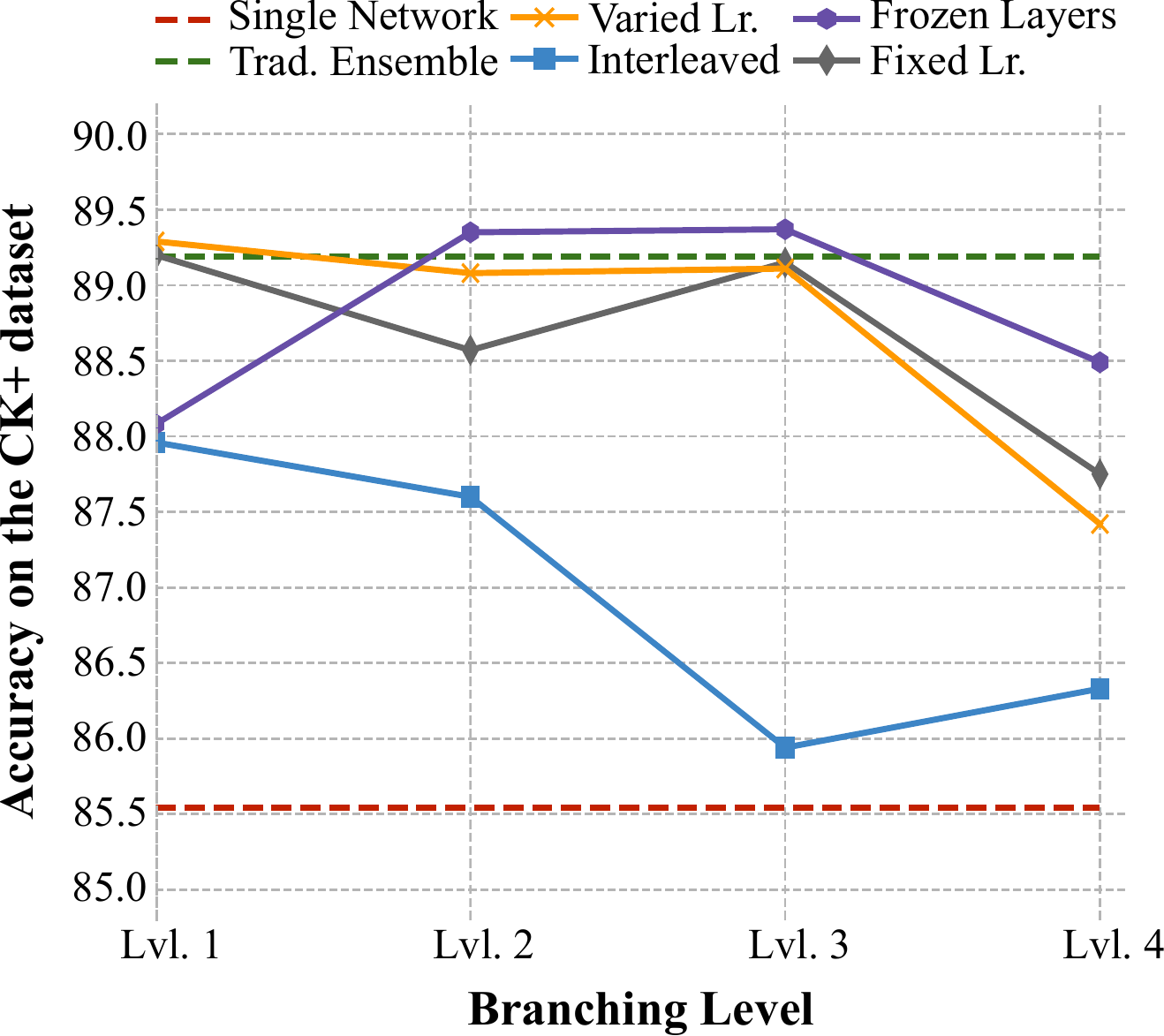}
	\caption{Accuracy on the Extended Cohn-Kanade dataset increasing branching level for different training strategies.}
	\label{fig:ck}
\end{figure}

The \textit{single network} was trained on four folds using stochastic gradient descent (SGD) to minimize the cross-entropy loss, whereas different training strategies using SGD were tested to build ensembles with complementary representations of the data. Given the small number of training samples, the \textit{traditional ensemble} was trained using bagging \cite{Breiman1996} due to its efficiency in dealing with the variance problem \cite{Dietterich2000}. Since we have four folds for training, we built an ensemble of four networks where each network was trained on three folds following a leave-one-fold-out scheme. The shared layers of ESRs allow us to test some variations of bagging. After adding a new convolutional branch to the ESR, the shared layers ($lr_{sl}$) and already trained branches ($lr_{tb}$) continue learning on additional data using (1) the same initial learning rate (\textit{fixed lr.}; $lr_{sl}=lr_{tb}=0.1$), (2) a smaller learning rate (\textit{varied lr.}; $lr_{sl}=0.1$ and $lr_{tb}=0.02$), or (3) not training at all (\textit{frozen layers}; $lr_{sl}=lr_{tb}=0.0$). We adopted a momentum factor of $0.9$ on SGD and a learning rate decay with a multiplicative factor of $0.5$ applied after every 250 epochs. Finally, we also included the \textit{interleaved} training strategy adopted in our previous work \cite{Siqueira2018a} in this experiment where all branches were trained iteratively on random mini-batches from the four folds. Data augmentation was randomly applied in all of the cases including brightness and contrast changes, horizontal flips, rotations up to 30 degrees, translations, and rescaling.

\begin{table}
	\caption{Test accuracy (\%) of the most accurate networks and baselines on CK+ and their number of parameters.}
	\centering
	\begin{tabular}{|| c | c | c ||}
		\hline
		Approach 						& {\scriptsize $\#$} 		& Accuracy						\\
		\hline \hline
		Single Network					& \boldmath$131.208$	& $85.5 \pm 3.5 \%$ 		\\
		Traditional Ensemble				& $524.832$	& $89.2 \pm 1.2 \%$ 					\\
		ESR-4 Lvl. 3					& $355.104$	& \boldmath$89.4 \pm 2.2 \%$ 			\\
		ESR-4 Lvl. 4					& $243.936$	& $88.5 \pm 3.8 \%$ 						\\
		\hline
	\end{tabular} 
	\label{tab:ckcategorical}
\end{table}
\begin{table}
	\caption{Paired \textit{t}-test (\textit{p}-values) to compare Single Network, Trad. Ensemble, ESR-4 Lvl. 3, and ESR-4 Lvl. 4 on CK+.}	
	\centering
	\begin{tabular}{c | c c c c }
									& TE							& \textbf{Lvl. 3}						& Lvl. 4 				\\
		\hline
		Single Network			& $0.004$ \cmark			& \boldmath$0.005$ \cmark		& $0.043$ \cmark	\\
		Trad. Ensemble (TE)		& $-$							& \boldmath$0.956$ \xmark			& $0.614$ \xmark		\\
		Lvl. 3						& $-$	 						& $-$									& $0.514$ \xmark		\\
	\end{tabular} 
	\label{tab:ckcategoricalttest}
\end{table}
\begin{figure*}[!t]
	\centering
	\includegraphics[width=1.0\textwidth]{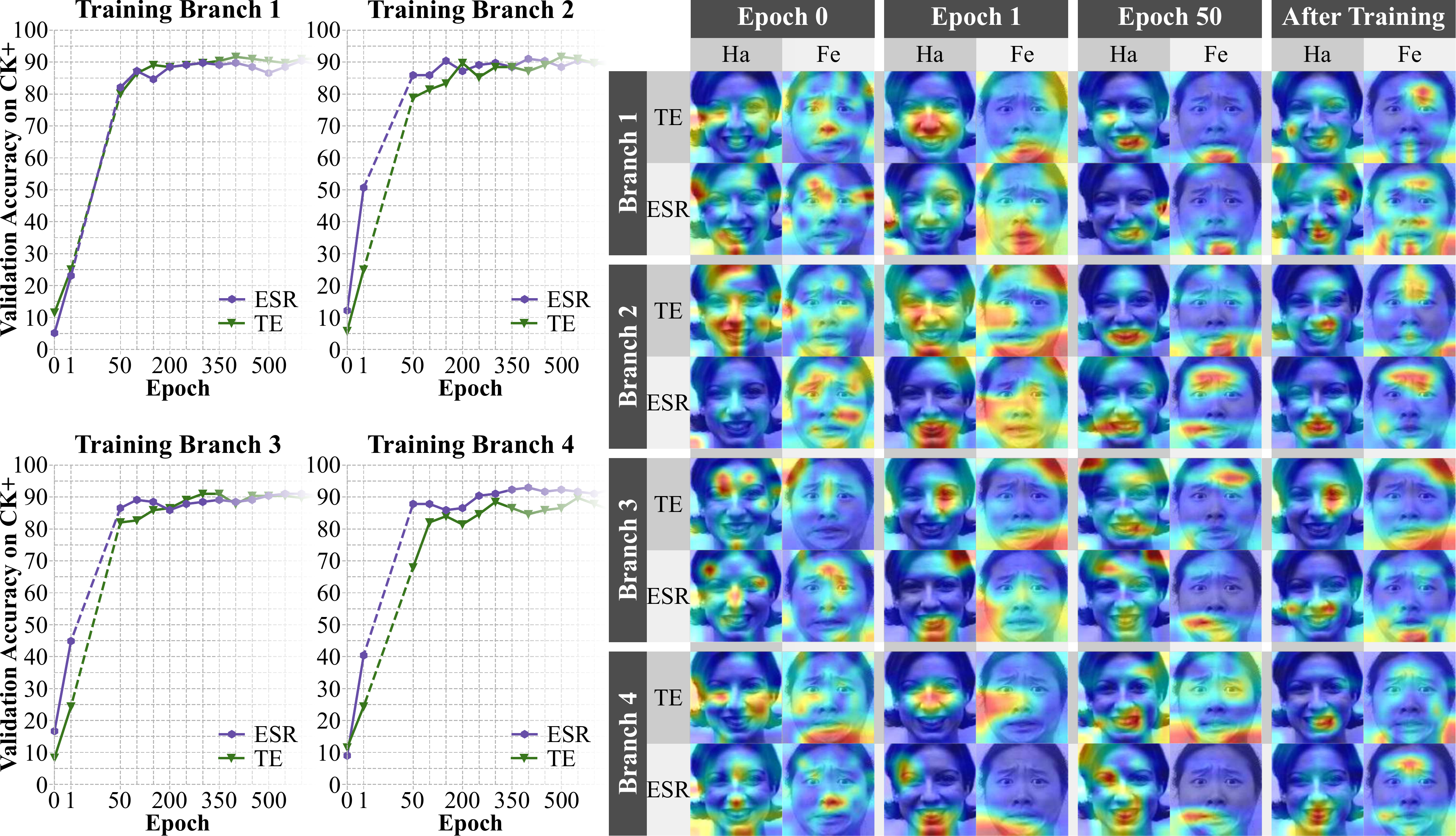}
	\caption{Comparison of the accuracy (\%) on the validation set of CK+ over epoch between the ensemble with shared representations and the traditional ensemble. On the right, Grad-CAM visualization at different training milestones including before any weight update (epoch 0). The saliency maps were generated using the jet colormap, where red regions indicate facial features that contributed the most to the high activation of the output neurons, in this example, happy or fear. Best viewed in color.}
	\label{fig:convergence}
\end{figure*}

Figure \ref{fig:ck} displays the mean accuracy on the CK+ test set with increasing branching level for every approach as well as the baselines (dashed lines). Consistent with ensemble literature, the ensemble methods achieved higher accuracies than the single network. The interleaved approach, however, demonstrated inferior performance among the ensembles. We believe the poorer performance might have been caused by low diversity in the ensemble. In interleaved training, diversity derives only from different starting points and different data augmentation executions on shuffled mini-batches. The accuracies obtained by the ESRs especially at level 3 were as high as the traditional ensemble method but the advantage is evident in the number of trainable parameters used by each approach, as shown in Table \ref{tab:ckcategorical}. ESRs need far less trainable parameters than traditional ensembles with a substantial decrease of $32\%$ at level 3 and $54\%$ at level 4, while achieving the same generalization power confirmed by the paired \textit{t}-test in Table \ref{tab:ckcategoricalttest}. Positive markers indicate statistically significant differences $(\textit{p} < 0.05$). The improvement in recognition performance is clear when compared with a single network. The high \textit{p}-value between the ESR with four branches at level 3 and TE indicates that the generalization abilities are equivalent while the redundancy and computational load are significantly reduced by ESR-4 Lvl. 3.

\subsection{Transfer Learning and Diversity Analysis}
Training time is an important factor when training deep neural networks, especially, ensembles of them. Figure \ref{fig:convergence} shows both how transfer learning in ESRs accelerates and guides the learning of new branches, as well as the diversity analysis of learned facial features. The graphs, on the left, compare the convergence of the ESR and TE over epoch for every branch, or network, added to the ensemble. The convergence curve follows the same pattern as the ensemble size increases in TEs since any new model is trained from scratch. On the other hand, the convergence speed of additional branches in ESRs increases due to the prior knowledge stored in the shared layers. Even after the first update, the ESR was already twice as accurate as TEs, and this gap was only closed around epoch 50.

These quantitative results suggest that the shared layers learned informative facial features of emotion concepts. To support our claim with visual evidence, we generated saliency maps with respect to the ESR and TE predictions at different training milestones using Grad-CAM. Note, on the right, that the learning progress of facial features advanced at the same pace while the ensemble size was one (both methods are identical). When training the second branch, however, the ESR already learned after the first update that the region around the mouth is relevant for recognizing happy facial expressions, whereas the TE took around 50 epochs to discover the same pattern.

In fact, AU-12 from the FAC must be visible on happy facial expressions in CK+ \cite{Lucey2010}. When the facial muscle underlying AU-12 (i.e., \textit{Zygomatic Major}) is fired, it pulls the corner of the lips up, producing a smile. The smile is one of the most discriminative and repetitive facial features presented in CK+ that distinguishes happy facial expressions among other categories. Fear, for instance, is categorized from the combination of more complex facial features, which would require more time for the ESR and TE to learn such features. Nevertheless, the ESR learned around epoch 50 that frowning is one of the features necessary to recognize fear. This appearance change is produced by the \textit{frontalis} muscle that covers the forehead and, when activated, can raise the inner brow (AU-1). The other feature is the lip stretcher coded by AU-20 and presented in fear facial expressions in CK+. In general, the TE needs more training epochs than ESR to learn informative facial features. Finally, note that the diversity of features of the ESR is as high as the TE. In the happy facial expression example, while branches 2 and 4 captured the smile after training, branch 3 focused on the nasolabial furrows and branch 1 captured the wrinkles in the outer eyes caused by raising the cheeks (AU-6).

\section{Training ESRs on Large-Scale Datasets}
Nowadays, training ensembles on large-scale datasets became impracticable for those who have limited computational resources because even a single deep neural network may take over a month of training using several GPUs in large data centers \cite{Chollet2017}. This section supports that ESRs are affordable for ensembling on large-scale datasets. Besides short training time and low computational cost, ESRs can reduce the remaining residual generalization error which led to higher accuracies than state-of-the-art methods in facial expression recognition benchmarks.

\subsection{Methodology}
Along with data, benchmark datasets usually provide standard experimental protocols and baseline results. AffectNet and FER+ have divided the dataset into training, validation, and test sets, and published them for the scientific community, except for the test set of the former. Meanwhile, researchers have utilized the validation set for evaluation and comparisons, as suggested by the AffectNet authors. We followed the same methodologies as the state-of-the-art methods for fair comparisons on both datasets. In experiments on AffectNet, the best inference performance on the validation set is reported, whereas the mean and standard deviation of the test accuracies after five trials are used as an evaluation metric for FER+ \cite{Barsoum2016}.

\subsection{Evaluation on the AffectNet Dataset}
As the body of features increases, the memory capacity of the neural network shall also increase to account for the higher volume of patterns. Therefore, the architecture used from this point is based on the previous ESR but with a few more convolutional layers, and batch normalization layers, as well as more convolutional filters per layer. In order to preserve the spatial information of the features, we adopted the same spatial reduction rate of the feature maps from the previous experiment by adding a max-pooling layer after every two convolutional layers. The ensemble of convolution branches begins in an equivalent spatial level to ESR-4 Lvl. 3, where the shape of the feature maps are similar, as depicted in Figure \ref{fig:architecture}.

\begin{table}
	\caption{Accuracy (\%) on AffectNet for discrete emotions and number of emotion labels used in the experiments.}
	\centering
	\begin{tabular}{|| c | c | c ||}
		\hline
		Approach																& {\scriptsize $\#$} 	&	Acc $\downarrow$	\\
		\hline \hline
		\textbf{ESR-9 (Our network)} 										& \boldmath$8$	& 	\boldmath$59.3 \%$	\\
		AlexNet-WL (Mollahosseini et al. \citeyear{Mollahosseini2019})	& 8 				& 	$58.0 \%$				\\
		VGGNet \cite{Hewitt2018}												& 8 				& 	$58.0 \%$				\\
		MobileNet \cite{Hewitt2018}											& 8 				& 	$56.0 \%$				\\
		AlexNet \cite{Hewitt2018}												& 8 				& 	$56.0 \%$				\\
		AlexNet-US (Mollahosseini et al. \citeyear{Mollahosseini2019})		& 8 				& 	$47.0 \%$				\\
		AlexNet-DS (Mollahosseini et al. \citeyear{Mollahosseini2019})		& 8 				& 	$40.0 \%$				\\
		\hline \hline
		gACNN \cite{Li2019} 													& 7 				& 	$58.8 \%$				\\
		IPA2LT \cite{Jiabei2018} 												& 7 				& 	$57.3 \%$				\\
		pACNN \cite{Li2019} 													& 7 				& 	$55.3 \%$				\\
		\hline
	\end{tabular} 
	\label{tab:affectcategorical}
\end{table}
\begin{figure}[ht]
	\centering
	\includegraphics[width=.47\textwidth]{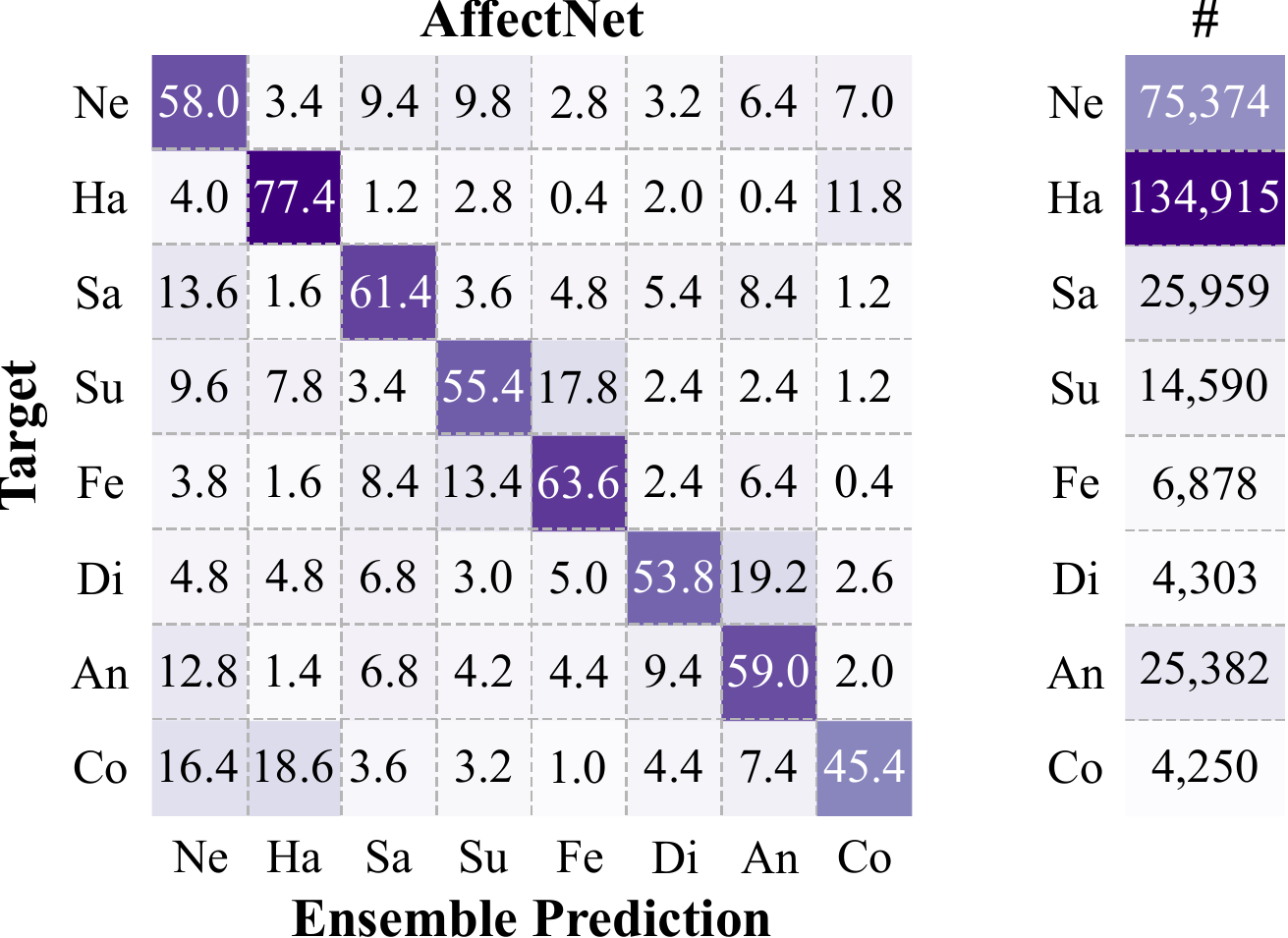}
	\caption{Normalized confusion matrix of the ensemble predictions on the AffectNet dataset and the emotion label distribution.}
	\label{fig:affectnet_cm}
\end{figure}

\textbf{Discrete emotion perception.} One of the challenges of facial expression recognition on AffectNet is the imbalance problem. We coped with this problem by training the branches of the ESR on balanced subsets from the whole training set containing up to 5000 samples of each emotion. Through an empirical analysis, subsets with fewer samples of each category resulted in lower performance, while more samples provided no significant gain in accuracy. The stochastic gradient descent was used to minimize the cross-entropy loss function with an initial learning rate of $0.1$, a momentum of $0.9$, and a learning rate decay with a multiplicative factor of $0.5$ applied after every 10 epochs. Convolutional branches were added to the ensemble until no significant gain in accuracy was achieved by the collective classification. Trained branches were continually updated on additional training data with a lower initial learning rate of $0.01$ for their adaptation to the representational changes in the shared layers.

Our results are reported in Table \ref{tab:affectcategorical} and Figure \ref{fig:affectnet_cm}. The ESR with 9 convolutional branches (ESR-9) achieved the highest recognition performance on AffectNet in comparison to state-of-the-art methods. It is important to note that no single branch in the ensemble achieved an accuracy higher than $58.0\%$, only the collective classification made by the ensemble reached $59.3\%$ of accuracy, which reveals that the remaining residual generalization error was reduced by the ESR. In the confusion matrix, we can see that ESR-9 was more accurate in the recognition of the happiness category but under-represented categories such as fear, disgust, and contempt were still well recognized given the disparity of the label distribution. Finally, even though \citeauthor{Li2019} (\citeyear{Li2019})'s ensemble has obtained an accuracy of $58.8\%$, the contempt category was removed from their experiments, which greatly reduced the complexity of the task since the chance of the network for learning undesired features decreased.

\begin{table}
	\caption{Root-mean-square error (RMSE) for arousal (aro) and valence (val) prediction on the AffectNet dataset.}
	\centering	
	\begin{tabular}{|| c | c | c ||}
		\hline
		\multirow{2}{*}{Approach} 									& 	\multicolumn{2}{c||}{RMSE}						\\
		\cline{2-3}
																		& 	Aro	 $\downarrow$		& 	Val	 				\\
		\hline \hline
		\textbf{ESR-9 (Our network)} 								& 	\boldmath$0.33$			& 	\boldmath$0.36$		\\
		VGGNet \cite{Hewitt2018}										& 	$0.37$					& 	$0.41$				\\
		MobileNet \cite{Hewitt2018}									& 	$0.38$					& 	$0.42$				\\
		AlexNet \cite{Hewitt2018}										& 	$0.39$					& 	$0.41$				\\				
		AlexNet (Mollahosseini et al. \citeyear{Mollahosseini2019})	& 	$0.41$					& 	$0.37$				\\
		VGG16-Based \cite{Lindt2019}								& 	$0.41$					& 	$0.45$				\\
		\hline
	\end{tabular}
	\label{tab:affectdimentional}
\end{table}

\textbf{Continuous affect perception.} Predicting arousal and valence levels of facial images in a continuous space is a complex task where disagreement levels between human annotators are usually higher than in discrete emotion annotations. Thus, we trained the ESR in a curriculum learning fashion \cite{Bengio2009}, where ESR-9, trained for discrete facial expression recognition, was fine-tuned for arousal and valence prediction. We assumed that some facial features learned by ESR-9 from the previous task would lead the network to learn faster and become more accurate for inferring affect concepts than training it from scratch. For example, a smile detector usually learned after a few training epochs, as shown in our experiments on the in-the-lab dataset, can be associated with positive arousal and positive valance levels. Instead of replacing the output layer of ESR-9 to account for arousal and valence predictions, we added two neurons on top of each branch, as shown in Figure \ref{fig:architecture}, and trained only the weights connected to those neurons. Since the relation of discrete emotions and continuous affect is non-linear, we applied the ReLU function to the second last layer that is related to discrete emotion concepts.

We followed the same training procedure as in our previous experiments where each branch is sequentially trained on a balanced subset with up to 5000 samples from each quadrant of the arousal and valence circumplex to reduce bias. However, since arousal and valence prediction in the continuous domain is a regression problem, we minimize the root-mean-square error using stochastic gradient descent with a momentum of $0.9$ and a learning rate of $0.01$. Trained branches were continually updated with a lower learning rate of $0.001$. The results are reported in Table \ref{tab:affectdimentional}. ESR-9 outperformed state-of-the-art methods based on established pre-trained deep neural networks for visual classification tasks with a significant margin on both of the arousal and valence dimensions by achieving $0.33$ and $0.36$ RMSEs, respectively. Moreover, since only the output layer was trained, ESR-9 can still perform discrete emotion perception which resulted in a great drop in computational load and redundancy. In comparison to \citeauthor{Mollahosseini2019} (\citeyear{Mollahosseini2019})'s approach which has approximately $180M$ parameters in total, ESR-9 has 9 times fewer parameters ($\approx20M$). In their work, three AlexNets were trained, one for each of the three facial expression perception problems. Finally, ESR-9 reached the performance of human experts in facial expression annotations which have a disagreement level of $0.36$ and $0.34$ RMSEs for arousal and valence prediction.

\subsection{Fine-tuning on the FER+ Dataset}
In our experiments on FER+, we rescaled the images from 48 x  48 pixels to 96 x 96 pixels and fine-tuned ESR-9 trained on AffectNet. Before any training on FER+, ESR-9 achieved a test accuracy of $57.92\%$, similar to its performance on AffectNet. This cross-dataset evaluation indicates that ESR-9 generalizes well to different data distributions. It is important to note that facial images from FER+ are gray-scale images with low resolution and are not as centralized as AffectNet's images. These aspects may deteriorate certain facial features in images relevant for emotion perception until a point where they can no longer be detected, as argued by \citeauthor{Tian2005} (\citeyear{Tian2005}). We fine-tuned ESR-9 using the stochastic gradient descent with a momentum of $0.9$, an initial learning rate of $0.1$, and a learning rate decay with a multiplicative factor of $0.75$ applied after every 10 epochs. The learning rate decay was increased due to the faster convergence of ESR-9 on FER+. Trained branches were continually updated with a lower initial learning rate of $0.02$.

\begin{figure}[ht]
	\centering
	\includegraphics[width=.47\textwidth]{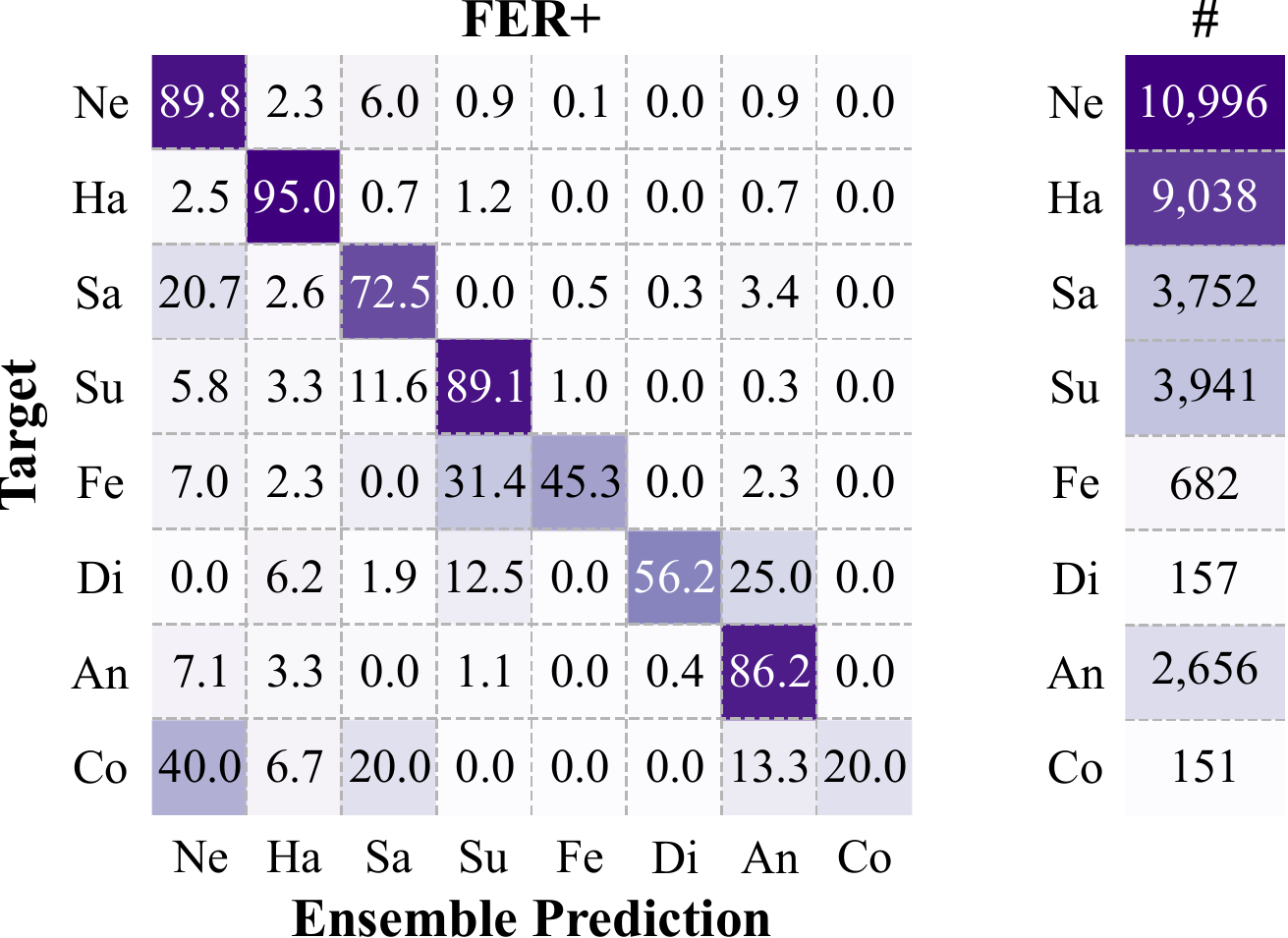}
	\caption{Normalized confusion matrix of the ensemble predictions on FER+ and the emotion label distribution.}
	\label{fig:ferplus_cm}
\end{figure}

\begin{figure}[ht]
	\centering
	\includegraphics[width=.47\textwidth]{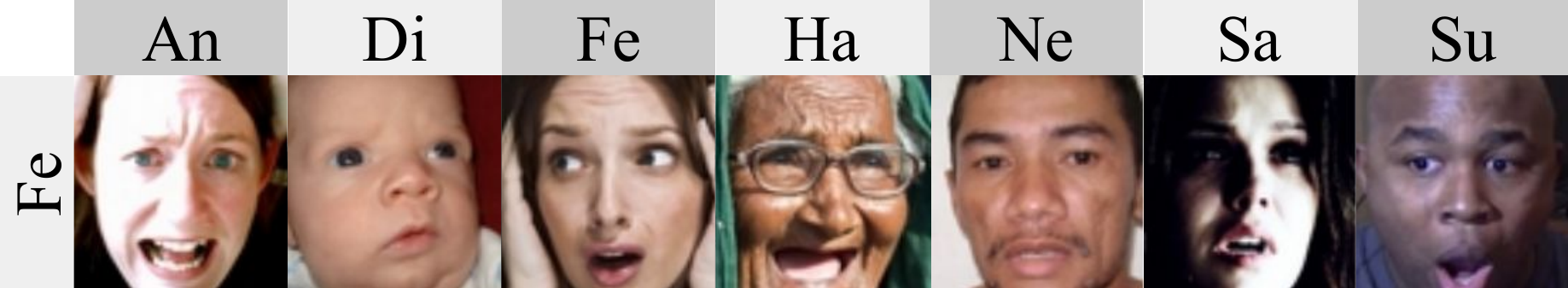}
	\caption{Subjective perception of facial expressions. Samples annotated as fear by one expert human annotator perceived differently by another expert. Adapted from \cite{Mollahosseini2019}.}
	\label{fig:subjectivity}
\end{figure}

After fine-tuning each branch sequentially on random subsets with up to 5000 training samples per emotion category on FER+, ESR-9 reached an average test accuracy of $87.153\%$ with a very low standard deviation of $0.097\%$, outperforming the current state-of-the-art method \cite{Barsoum2016}. Our results are reported in Table \ref{tab:ferplus} and Figure \ref{fig:ferplus_cm}. Also, ESR-9 generalized reasonably well to under-represented categories. When compared to \citeauthor{Barsoum2016} (\citeyear{Barsoum2016})'s approach, for instance, ESR-9 correctly recognized $20.0\%$ of the contempt test samples and $56.2\%$ of the disgust test samples, while their approach recognized only $4.17\%$, and $26.32\%$ respectively. The bias towards neutral classifications was also reduced in almost all categories. While our approach misclassified $40\%$ of contempt samples as neutral and had no misclassification of disgust samples as neutral, their approach misclassified $54.17\%$ of contempt and $10.53\%$ of disgust samples. The bias problem in facial expression recognition is not solely related to the unbalance problem, but also to the inherent subjectivity of emotion perception on faces where humans may perceive different emotions in the same facial expression as illustrated in Figure \ref{fig:subjectivity} \cite{Barrett2017,Mollahosseini2019}.

It is relevant to mention that we extensively investigated the effects of varying the maximum number of training samples for each emotion category on FER+. When trained with lower upper bounds, ESRs increased recognition on under-represented categories but the overall accuracy decreased. If the upper bound is too high, the diversity of the ensemble decreased. In this experiment, an upper-bound of 5000 samples for each category was the ``optimal'' value to achieve high overall accuracy and a relatively high correct classification of under-represented categories. Finally, our findings suggest that our approach to ESRs is an important contribution to alleviate the bias problem in machine learning.

\begin{table}
	\caption{Mean and standard deviation of the test accuracy on FER+. Some authors only reported the best accuracy.}
	\centering
	\begin{tabular}{|| c | c ||}
		\hline
		Approach								& 	Acc	 $\downarrow$ 				\\
		\hline \hline
		\textbf{ESR-9 (Our network)} 		& 	\boldmath$87.15 \pm 0.1 \%$	\\
		SHCNN \cite{Miao2019} 				& 	$86.54 \%$						\\
		VGG16-PLD \cite{Barsoum2016} 		& 	$84.99 \pm 0.37 \%$				\\
		VGG16-CEL \cite{Barsoum2016} 		& 	$84.72 \pm 0.24 \%$			\\
		TFE-JL \cite{Li2018} 					& 	$84.3 \%$							\\
		VGG16-ML \cite{Barsoum2016} 		& 	$83.97 \pm 0.36 \%$				\\
		VGG16-MV \cite{Barsoum2016} 		& 	$83.85 \pm 0.63 \%$			\\
		ResNet18 + FC \cite{Li2018}			& 	$83.4 \%$							\\
		ResNet18 \cite{Li2018} 				& 	$83.1 \%$							\\
		\hline
	\end{tabular} 
	\label{tab:ferplus}
\end{table}

\section{Conclusions}
Referring to Minsky at the beginning of this paper, one may think that single deep neural networks trained on large-scale datasets are enough to build rich, robust and highly accurate perceptual models. However, in certain domains where label distribution is unbalanced, for instance, those networks tend to become highly biased to the most representative categories. We demonstrated that ensembles with shared representations cope with this problem by training ``many resources'' (i. e., convolutional branches) on balanced subsets from the training data. Together, through the collective classification made by the ensemble, ESRs outperformed state-of-the-art deep neural networks on AffectNet and FER+ with low redundancy and an efficient transfer learning mechanism from the shared layers. Moreover, we showed that the branching level directly impacts ensemble diversity, generalization, and computational load.

Artificial neural networks, when trained under continual learning settings, typically suffer from a phenomenon called \textit{catastrophic forgetting}. Correct classified samples become misclassified when the network is continually trained on a different data distribution due to its inability to keep learned information. The same phenomenon occurs when training additional branches in ESRs having a direct impact on the generalization performance. To address the effects of catastrophic forgetting on ESRs, learning rates of the trained branches and shared layers should be carefully defined. High differences in learning rates may cause trained branches to forget learned information, whereas similar learning rates may foster co-adaptation between branches and decrease ensemble diversity. In the future, we will investigate approaches to overcome catastrophic forgetting in ensembles with shared representations.

Despite reaching human-level performance in facial expression recognition on AffectNet, human-level affect inference under real-world conditions is far to be reached. To do so, computational models closer to recent findings that are changing and enhancing our understanding of emotions under the theory of psychological construction \cite{Barrett2015} should be developed. It is important to take into consideration not only cross-modal learning of emotional expressions but also temporal and contextual information during emotional episodes. As the next step, we will implement a model closer to the theory of constructed emotion \cite{Barrett2017} by adopting ESR-9's high-level representations as ``proto concepts'' of facial expressions to guide learning of emotion concepts in a hybrid neural system based on an intermediate view between empiricism and nativism of the cognition theory \cite{Ullman2019}.

\section{Acknowledgments}
This work has received funding from the European Union's Horizon 2020 research and innovation program under the Marie Sk\l{}odowska-Curie grant agreement No. 721619 for the SOCRATES project. The authors thank Prof. Dr. Thomas Hellstr\"{o}m for his insightful questions that motivated the development of this paper.

\bibliographystyle{aaai}
\bibliography{ref}

\end{document}